\documentclass[sigconf]{acmart} 

\AtBeginDocument{%
  \providecommand\BibTeX{{%
    \normalfont B\kern-0.5em{\scshape i\kern-0.25em b}\kern-0.8em\TeX}}}

\copyrightyear{2020}
\acmYear{2020}
\setcopyright{acmcopyright}
\acmConference[MM '20] {28th ACM International Conference on Multimedia}{October 12--16, 2020}{Seattle, WA, USA.}
\acmBooktitle{28th ACM International Conference on Multimedia (MM '20), October 12--16, 2020, Seattle, WA, USA.}
\acmPrice{15.00}
\acmDOI{10.1145/3394171.3413604}
\acmISBN{978-1-4503-7988-5/20/10}

\usepackage{graphicx}
\usepackage{multirow}
\usepackage{tabularx} 
\usepackage{caption}
\usepackage{xspace}
\usepackage{booktabs}
\usepackage{subfigure}
\usepackage{amsmath,amssymb} 
\usepackage{enumitem,xcolor}

\def \ie {{i.e.}\xspace}
\def \eg {{e.g.}\xspace}

\def \etal {{et al.}\xspace}
\def \vs {{v.s.}\xspace}
\newcommand{\miaojing}[1]{#1}

\newcommand{\teddy}[1]{{#1}}

\newcommand{\para}[1]{\noindent \textbf{#1}}

\usepackage{amsmath,epsfig}
\usepackage{diagbox}
 \usepackage{tkz-graph} 

\acmSubmissionID{1200}

\begin{document}
\author{Wenqing Liu}
\authornote{Wenqing Liu and Miaojing Shi contributed equally; Li Li is the corresponding author.}
\affiliation{College of Electronic and Information Engineering\\ Tongji University, Shanghai, China}
\email{liuwenqing@tongji.edu.cn}

\author{Miaojing Shi}
\affiliation{King's College London \\ London, UK}
\email{miaojing.shi@kcl.ac.uk}

\author{Teddy Furon}
\affiliation{Univ. Rennes, Inria, CNRS, IRISA \\ Rennes, France}
\email{teddy.furon@inria.fr}

\author{Li Li}
\affiliation{College of Electronic and Information Engineering\\
Institute of Intelligent Science and Technology\\
Tongji University, lili@tongji.edu.cn}

\renewcommand{\shortauthors}{Liu, et al.}

\title{Defending Adversarial Examples via DNN Bottleneck Reinforcement}



\begin{abstract}
  This paper presents a DNN bottleneck reinforcement scheme to alleviate the vulnerability of Deep Neural Networks (DNN) against adversarial attacks.
Typical DNN classifiers encode the input image into a compressed latent representation more suitable for inference.
This information bottleneck makes a trade-off between the image-specific structure and class-specific information in an image.
By reinforcing the former while maintaining the latter, any redundant information, be it adversarial or not, should be removed from the latent representation.
Hence, this paper proposes to jointly train an auto-encoder (AE) sharing the same encoding weights with the visual classifier. In order to reinforce the information bottleneck, we introduce the multi-scale low-pass objective and multi-scale high-frequency communication for better frequency steering in the network. Unlike existing approaches, \miaojing{our scheme is the first reforming defense per se} which keeps the classifier structure untouched without appending any pre-processing head and is trained with clean images only.
Extensive experiments on MNIST, CIFAR-10 and ImageNet demonstrate the strong defense of our method against various adversarial attacks. 
\end{abstract}




\maketitle

	\begin{figure}
\centering
\includegraphics[width=0.95\columnwidth]{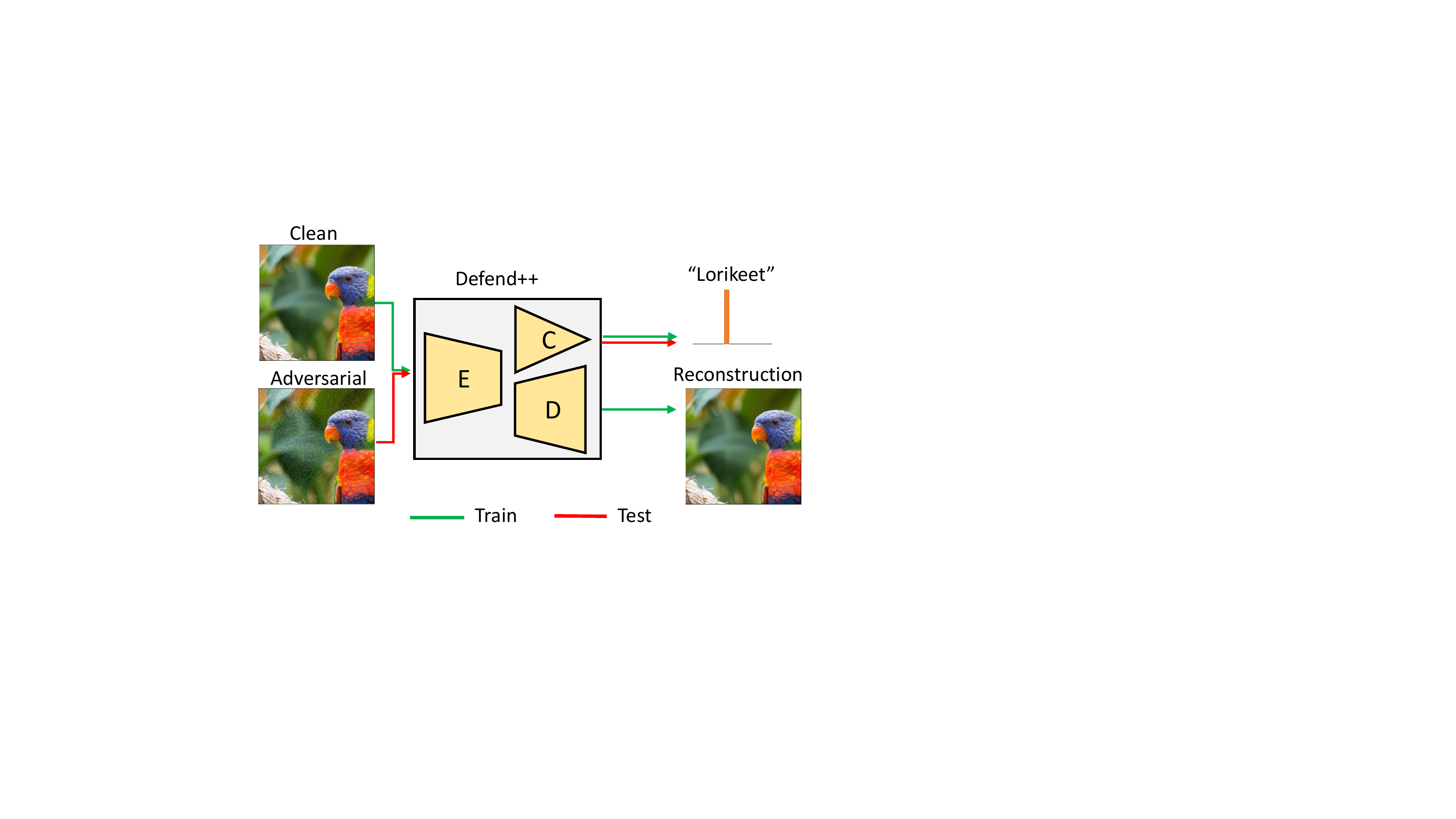}
\captionof{figure}{Illustration of our Defend++: defending adversarial examples via DNN bottleneck reinforcement. The structure of a classifier is untouched but split into two parts (E+C). A specifically designed D-ecoder is appended during the training to form an auto-encoder (E+D).The structure is trained on clean images only to jointly minimize the classification (E+C) and the reconstruction (E+D) losses.}
\label{Fig:motivation}
\vspace{-2mm}
\end{figure}

\section{Introduction}
Deep learning achieves excellent performance in various multimedia and computer vision tasks~\cite{krizhevsky2012nips,shi2019cvpr,yang2019training}, but it has weaknesses.
Researchers have observed that specially crafted perturbations to images/videos lead to total failure of visual recognition~\cite{szegedy2014iclr,goodfellow2015iclr,jiang2019mm}.
Surprisingly, these perturbations have very small amplitude so they are barely perceptible.
It means that in the image space, natural images lie close to the class boundary and are easy to be perturbed.

The literature on adversarial images considers two scenarios.
In the while-box scenario, the attacker knows everything about the classifier.
The backpropagation computes the gradient of this classifier's loss w.r.t. the image pixel values~\cite{zhao2018mm}.
It indicates the infinitesimal perturbation in the image space dragging the image closer to the class boundary~\cite{szegedy2014iclr}.
This operation can be iterated until the image crosses the boundary so that the model prediction is wrong~\cite{kurakin2017iclr}.
Adversarial examples also happen to transfer across different models.
In the black-box scenario, the attacker is oblivious to the target model internals.
Yet, by observing the target's outputs for many inputs, the attacker can train his own model mimicking the target.
By leveraging transferability, a white-box attack on his own classifier often deludes other black-box models~\cite{meng2017sigsac}. 

Many defenses have been proposed but they resort to few different mechanisms.
One idea coined as \emph{gradient masking}~\cite{papernot2016sp,buckman2018iclr} targets the core of white-box attacks,
the computation of gradient, by blocking the backpropagations.
Yet, gradient masking has been shown to give a false sense of security~\cite{carlini2017sp}:
this defense is not efficient against black-box attacks, which are free from any gradient computation on the target model. 

\emph{Adversarial training} robustifies the classifier
by pushing the class boundary away so that adversarial images lie in the true class region~\cite{szegedy2014iclr,goodfellow2015iclr}.
The fine-tuning of model parameters is time-consuming.
This is why adversarial training usually considers fast but simple attacks.
Its strength against more elaborated attacks remains questionable.

A good defense should not be dedicated to any specific attack or mechanism, which we conceptualize as its  \emph{universality}. 
The class boundaries learned by the classifier are valid over the manifold of natural images, while perturbations look like noise and perturbed images are not lying on the manifold of natural images; the inference of the classifier can be awfully wrong when probing images off this manifold.
This motivates pre-processing the input image in front of the classifier~\cite{meng2017sigsac}.   
The goal is to reform suspicious images, \ie to project them back onto the manifold of natural images~\cite{bakhti:hal-02349625}.
The most popular choice for this front-end pre-processing is an auto-encoder (AE)~\cite{gu2014arxiv,liao2018cvpr,bakhti:hal-02349625}. It is usually trained with adversarial examples forged by many attacks. Few-shot learning can be adopted to accelerate this process~\cite{ma2019mm}. Nevertheless, 
training it with clean images can strongly enforce the universality of the defense.
This is recently achieved in~\cite{jia2019cvpr}.


This paper proposes to defend adversarial examples via DNN bottleneck reinforcement, which we call it Defend++. It pertains to the defense trend of universality yet is not a separated front-end input reformer. 
Instead, our idea is to merge the reformer (\ie AE) with the early layers of the classifier so that both functionalities are jointly trained.  Once training is finished, the inference of the classifier is disjointed from the AE (see Fig.~\ref{Fig:motivation}). 
Our architecture can be seen in two ways:
it is a classifier working directly over the internal latent space representation of the AE (instead of being connected at its output).
From another point of view, it is a visual classifier whose first layers are jointly trained for both classification and reconstruction. 

Our motivation \teddy{is based on the following view of a classifier. A DNN} encodes an image into a compressed latent representation compressing the image-specific structure (\ie edges, corners, and regions) to concentrate the class-specific information (\ie object size, location, and common traits of the class).
It has been theoretically justified as necessary for better classification by the concept of information bottleneck~\cite{michael2018on}. 
If we could reinforce the image-specific structure while maintaining the class-specific information, any redundant information, adversary or not, would be removed without harming the prediction. This information bottleneck reinforcement 
has to take place inside the network.

\teddy{Due to the fact that adversarial perturbations happen to be high frequency in nature, we reinforce the information bottleneck with frequency steering in the AE. We introduce the 1) multi-scale lowpass objective to gradually reconstruct the image from low-level structure to high-level details; \miaojing{2) multi-scale high-frequency communication} to  
separate the prediction of high-frequency bands from lowpass images and connect them to different parts of the encoder for high-level communications.} To achieve the universality, the training of Defend++ deals with clean images only.
We do not alter the original classifier structure and no pre-processing is placed up front. Details of the classifier and training strategy are public.
Our method is both attack agnostic and scenario agnostic. 

The contribution of Defend++ is three-fold: 
\begin{itemize}[topsep=2pt,parsep=0pt,partopsep=2pt]
    \item We improve the universal defensive ability of the classifier by jointly learning an AE with it sharing the same encoding weights and on clean images only.
    \item \miaojing{We reinforce the network information bottleneck by introducing the multi-scale low-pass objective and multi-scale high-frequency communication for the AE. }
   \item We conduct extensive experiments on MNIST, CIFAR-10 and ImageNet benchmarks and demonstrate strong defense of our Defend++ against various adversarial attacks. 
\end{itemize}

To our knowledge, Defend++ is the first \teddy{reforming} defense that requires neither modification on the architecture of classification net, nor adversarial data for training. 

	\section{Related works}\label{Sec:RelatedWorks}
This section surveys adversarial learning from the attack and defense perspective, respectively. We mainly review white-box attacks as they are the hardest to defend. 

\subsection{Attack methods}
All white-box attacks benefit from the cheap computation of the gradient direction of the loss thanks to the backpropagation.
In the initial discovery of adversarial examples in~\cite{szegedy2014iclr}, the Fast Gradient Sign Method (FGSM) minimizes a first-order approximation of the loss over the ball of $\ell_\infty$ norm $\epsilon$. Kurakin~\etal~\cite{kurakin2017iclr} further introduce a basic iterative method (BIM) by repeating FGSM for several steps.
Carlini-Wagner~\cite{carlini2017sp} looks for the adversarial perturbation of minimal $\ell_2$ norm. The solution is given by a Lagrangian formulation that combines the gradients of the loss and the Euclidean distortion.
Yet, C\&W requires a high number of iterations as several Lagrangian multiplier values are tested. 
Recently, Rony~\etal~\cite{rony2019cvpr} introduce an efficient attack quickly finding an adversarial image and then refining it to minimize its distortion.
This is done by decoupling the direction and norm (DDN) of the perturbations.

\begin{figure}[t]
	\centering
	\includegraphics[width=0.95\columnwidth]{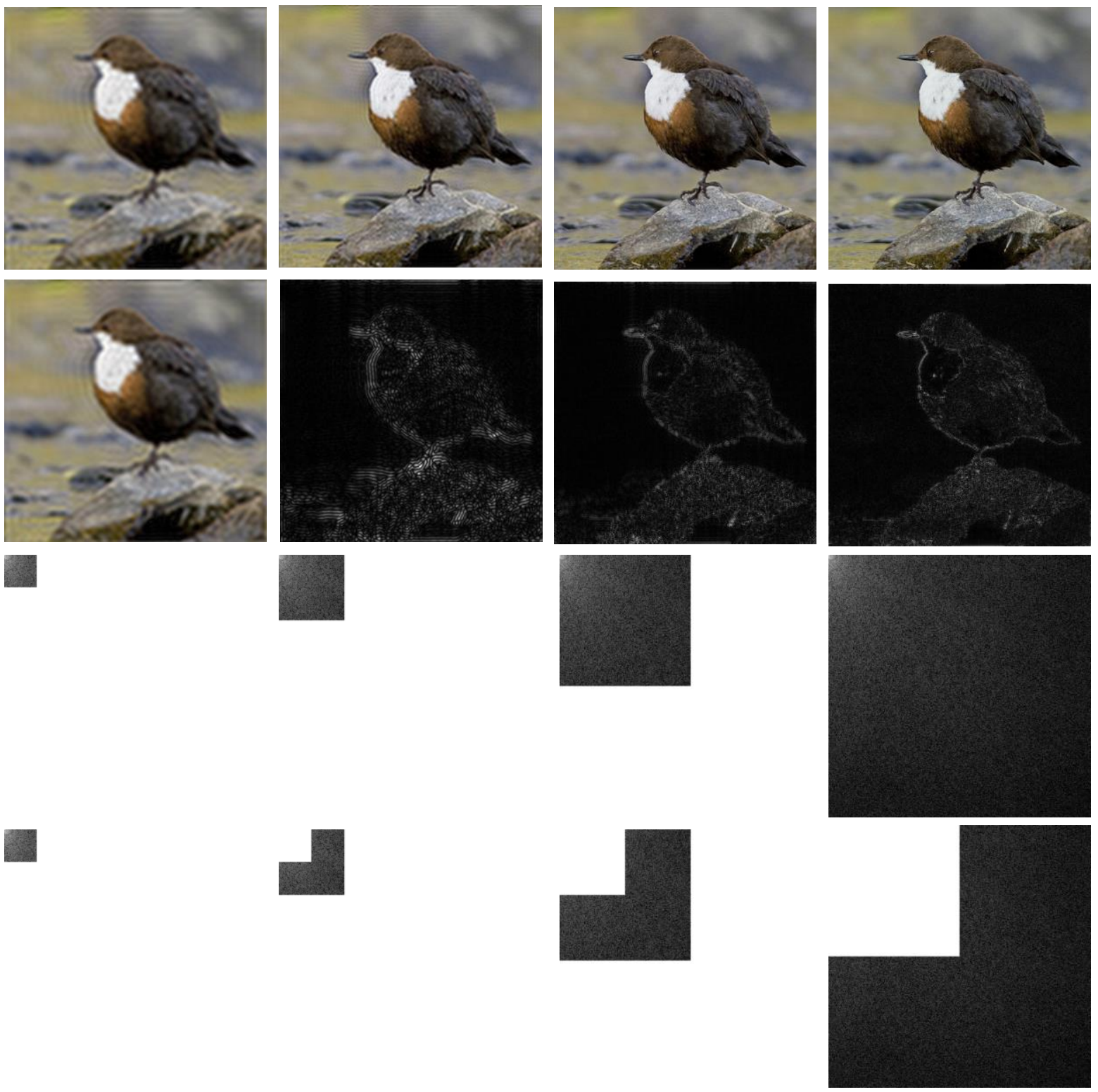}
	\caption{ First and third row: lowpass filtered images and corresponding frequency maps via DCT. Second and fourth row: \miaojing{Pyramid} images and corresponding frequency maps via DCT.
	}
	\label{Fig:dct}
	\vspace{-0.2cm}
\end{figure}

\subsection{Defense methods}
The first counter-attack, \emph{gradient masking}, blocks the backpropagation, which is at the core of any white-box attack, by introducing some randomization~\cite{taran2019cvpr}, smooth labels~\cite{papernot2016sp,buckman2018iclr}, soft feature selections~\cite{hua2019mm}, and regularizers~\cite{nayebi2017arxiv} to make
the model output less sensitive to the perturbation on input. The classification structure is often modified in these works~\cite{hua2019mm,taran2019cvpr}.  
Also, there exists a vulnerability against non gradient-based attacks (black-box) where the attacker is free from differentiating proxies of these methods
to circumvent the defense. This is outlined in~\cite{carlini2017sp}.

\emph{Adversarial training} is a promising idea where a network is re-trained with adversarial images. It performs well on small datasets such as MNIST and CIFAR, but decreases accuracy on large-scale datasets like ImageNet~\cite{kurakin2017iclr}. It is much costly.
An attack has to be mounted against a model, anytime that model is updated, new adversarial images must be generated.
This is the reason why adversarial training only uses fast attacks like FGSM which are not very powerful. The status may change in the future with the invention of fast and efficient attacks like DDN~\cite{rony2019cvpr}.

\teddy{A third idea is to detect adversarial images~\cite{Lu_2017_ICCV, 8237877}. This amounts to add the extra class ``suspicious images" as a possible output of the classifier. The attack is made harder as it must delude both the detector and the classifier. Training usually needs adversarial examples to minimize the false negative detection rate. This questions their ability to detect new attacks
~\cite{10.1145/3128572.3140444}.
As far as we know,~\cite{Ma2019NICDA} is the only detector trained on clean images only.}

A \teddy{fourth} idea is to utilize a pre-processing head in charge of filtering out any perturbations if existed. This is usually called a reformer.
Dziugaite ~\etal~\cite{dziugaite2016arxiv} report that JPEG compression can reverse small adversarial perturbations. Several variants, such as feature squeezing~\cite{xu2017arxiv}, pixel defend~\cite{song2017arxiv} and distillation~\cite{liu2019cvpr}, are further proposed to mitigate adversarial attacks.
Meng~\etal~\cite{meng2017sigsac} see the reformer as a projection of the input onto the manifold of clean data.
Gu~\etal~\cite{gu2014arxiv} were the first to propose an AE as a reformer. Liao~\etal~\cite{liao2018cvpr} introduce a high-level representation guided denoiser within the AE.
Yet, these advanced reformers mostly need training with adversarial examples, and the same question as for adversarial training \teddy{and detection} arises: how do they perform against attacks not encompassed in their training?

The recent paper~\cite{jia2019cvpr} shuts down the concern by proposing an efficient reformer (AE) that is trained without adversarial images. Its reformer is still separated from the classifier, and is carefully designed to compress over image bits with additional Gaussian noises.  Our AE instead shares the encoding part with the classification network and is jointly trained with the classifier without any additional noises.  
Moreover,~\cite{jia2019cvpr} assumes that the attacker knows the classifier but not the reformer. Ours is however totally open to attacker. We believe a complete white-box scenario makes more sense as nothing prevents the attacker to reproduce the reformer.


	\section{Method}\label{Sec:Method}

 \begin{figure}[t]
	\centering
	\includegraphics[width=0.95\columnwidth]{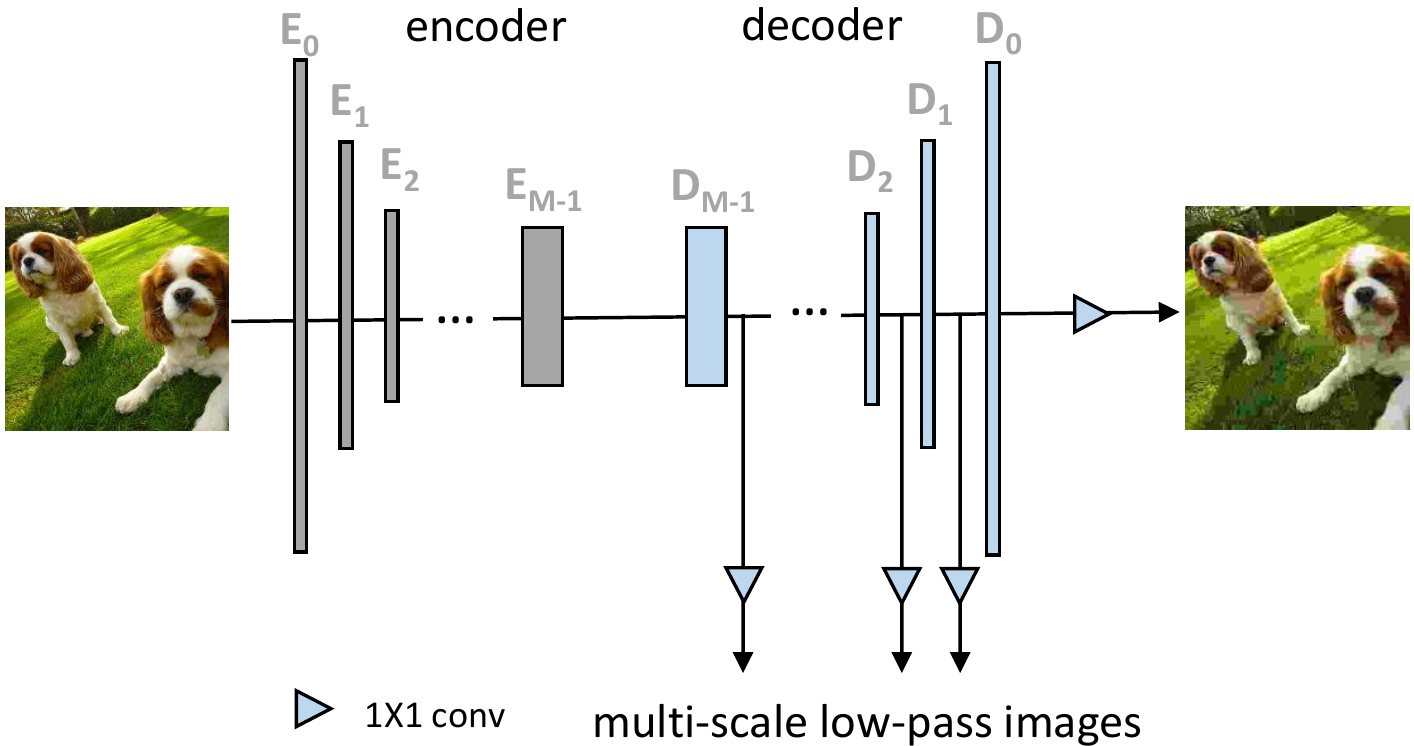}
	\caption{A simple auto-encoder (AE) with hourglass structure. We use $E$ and $D$ to denote the encoding and decoding block, respectively: $E_{M-1}$ ($D_{M-1}$) signifies a stack of encoding (decoding) blocks, while the others are with single block.     
	}
	\label{Fig:AE}
	\vspace{-0.3cm}
\end{figure}

 \begin{figure*}[t]
	\centering
	\includegraphics[width=0.85\textwidth]{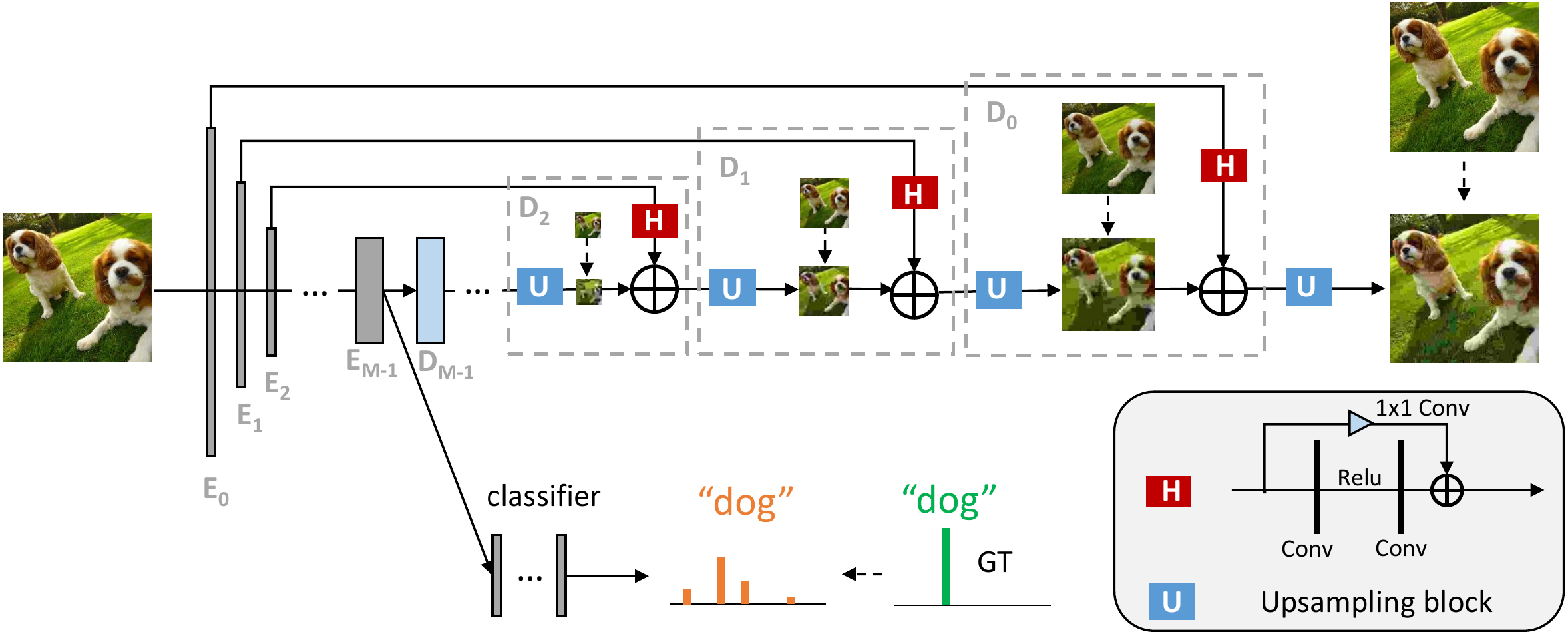}
	\caption{ \teddy{Overview of the reconstruction part of Defend++. The original image is low-pass filtered via DCT to create multi-scale ground truth for the decoder learning. Each decoding block $D$ is formed with $U$ and $H$ modules just like in a Laplace pyramid to separately tune the importance of each frequency band.}
	}
	\label{Fig:overview}
	\vspace{-0.3cm}
\end{figure*}

\subsection{Problem definition}
Given a visual recognition task, \ie classification,  the target of this work is to learn a robust model that generalizes on a wide range of clean images and defenses against various adversarial examples. 
In our setting, the model defensive ability can only be improved during classification training without whistles and bells.  This is  very challenging, but any effective method derived from this setting can be especially helpful. 

\subsection{Motivation}\label{Sec:motivation}
A typical classification network consists of multiple convolutional layers and several fully connected (fc) layers. The series of convolutional layers is regarded as an encoder producing a latent representation of the image, while fc layers act as a classifier making prediction upon it.
\teddy{This partition is very schematic and indeed any split of the network in two can be loosely seen as an encoder followed by a classifier.}

This latent representation contains two-fold information: the image structural information, \ie edges, corners and regions, that are most visually distinguishable; the class-specific information, \ie object size, location and appearance, that are most similar to others of the same class. There exist competitions between the two forces due to the nature of classification training. Intuitively, if we could reinforce the first fold while retaining the second fold information, any redundant information should be removed from the image representation and does no harm to the network prediction.

Referring to the adversary generation process, the imperceptible perturbation can be regarded as additional noise with some particular structures. It does not affect the original structure of the image but rather reflects the redundant information of the image~\cite{jia2019cvpr}. A common approach to remove image redundancy is to train an auto-encoder (AE). Previous methods employ AE for adversarial defense~\cite{gu2014arxiv,liao2018cvpr,jia2019cvpr} as an independent network for input transformation.
In our Defend++, the AE is jointly learnt with the classification net on clean images only.
Another difference is that we do not modify the architecture of the classifier once it is given. The encoder of the AE is indeed composed by the first layers of the classification net. In other words, the weights of these layers are learnt by merging both classification and reconstruction losses. 

\subsection{DNN bottleneck reinforcement}\label{Sec:joint}
AEs are neural networks that aim to copy their inputs to their outputs. They work by compressing the input into a latent-space representation, then reconstructing the output from this representation. The former part is conceptualized as an encoder removing the redundant information from the image while the latter part is a decoder reconstructing the image. The encoder consists of a set of convolutional layers downsampling the image while the decoder is a set of deconvolutional layers upsampling the image.

In order to distill the encoded knowledge between the classification net and the AE, we share the encoder of the AE with that of the classification net. Notice that when the classification net is too deep or too shallow, we can cut or extend the encoding blocks of the network considered as the encoder to an appropriate length. The decoder design is an open problem: a straightforward way is to employ the so-called hourglass structure~\cite{ronneberger2015miccai} where each decoding block is a mirrored version of the encoding block chaining the layers in reverse order (see Fig.~\ref{Fig:AE}). Here we use encoding/decoding blocks to refer to a stack of layers in the network where the input is only downsampled/upsampled once with a factor of 2 in this block. For instance, given a classifier ResNet-18, its encoding block normally consists of two/three residual blocks where the stride of the first convolutional layer in the block is set to 2 to downsample the image; a naive decoding block (\eg in hourglass structure) reveres the encoding block and replaces its last convolutional layer with a deconvolutional layer of stride 2 to upsample the image.

\teddy{Hourglass is a very simple AE. Below we first introduce a multi-scale lowpass objective to specifically improve its compressing ability against adversarial/noisy perturbations.The encoder in Fig.~\ref{Fig:AE} is only connected to the decoder through the network bottleneck, which is weak. While the encoder is inherited from the classification net and is untouched in our setting, we are free to design the decoder. 
In order to better steer the image frequency in encoding blocks, we further reform the decoding blocks \miaojing{into a pyramid structure} which separates the high-frequency band from each decoding output and connects it to an encoding block for direct communication.}   

\para{Multi-scale low-pass objective.}
Image noises including adversarial perturbations happen to be high frequency in nature.  
Hence, we believe each frequency band deserves a proper treatment. Inspired by \teddy{papers~\cite{dziugaite2016arxiv,xu2017arxiv} filtering out adversarial perturbations in certain extent thanks to the Discrete Cosine Transform (DCT), we propose a multi-scale low-pass objective to reconstruct the image gradually from low-level structure to the high-level details.}

Formally, let $F$ represent the frequency map of an image $I$ via the full DCT transform, $F = \text{DCT}(I)$. $F$ has the same size $W \times H$ with $I$.
A lowpass filtered version is generated by removing high frequency coefficients (see Fig.~\ref{Fig:dct} first and third rows).
The origin starts from the top-left corner of the map where the DC component lies in.
Suppose that the decomposition has $M$ levels with a scale of 2.
For the $m$-th level, $m\in\{0,1,\ldots,M-1\}$, the frequency map $F^m$ has size $W_m = \frac{W}{2^m}$ and  $H_m = \frac{H}{2^m}$ and coefficients $F_{wh}^m = F_{wh},\forall (w,h)\in \{1,W_m\}\times\{1,H_m\}$.
The lowpass filtered images $ I^0, I^1,..., I^{M-1}$ are obtained by applying inverse DCT on the $M$ lowpass frequency maps. Note that $I^0$ is indeed the original image $ I$. 

Compared to using linear interpolation to generate multi-scale ground truth, the proposed way provides a better feature steering to reconstruct different levels of details.


\para{Multi-scale high-frequency communication.} \teddy{
Following the notations above, we use $\tilde I^0$,..., $\tilde I^{M-1}$ to denote the reconstructed output corresponding to the multi-scale lowpass ground truth $I^0$,..., $I^{M-1}$. In the hourglass structure, the decoding blocks $D_{M-1}$,..., $D_{0}$ are chained in a reverse order of the encoding blocks $E_{0}$,..., $E_{M-1}$ with deconvolutional layers. 
Notice $D_{M-1}$
denotes a stack of multiple decoding blocks to decode the compressed representation at the bottleneck layer into $\tilde I^{M-1}$. This image has the smallest size and contains the low-frequency part; no ground truth is associated smaller than that. The rest $D_m$ denotes single decoding block. Similar interpretation is for $E_{M-1}$ and $E_m$.
The output of a given decoding block $D_m$ can thus be written as $\tilde I^m = D_m(D_{m+1}...(\tilde I^{M-1}))$. The decoding ability of the AE is enhanced with multi-scale lowpass objective by gradually restoring the high-frequency part of the image into the output (see Fig.~\ref{Fig:AE}). Notwithstanding, our ultimate goal is to improve the encoding ability of the AE to remove high-frequency adversarial/noisy perturbations.} Below we introduce our network reformation inspired by Laplacian pyramid. 

\teddy{A Laplace pyramid represents an image by the decomposition $LP(I) = [P^0,..., P^m,...,I^{M-1}]$, where $P^0,..., P^m$
are the detail layers (high-frequency bands) of the original image and $I^{M-1}$ the lowest resolution of the image (see Fig.~\ref{Fig:dct} second and fourth rows). The high frequency bands $P^0,..., P^m$ are disjointed from the lowpass images $I^0,..., I^m$ in the following  way:
\begin{equation}\label{eq:Lap}
    P^m = I^m  - U(I^{m+1})
\end{equation}
where $U$ is an upsampling operator.
}


\miaojing{Applying this pyramid to the decoder, each decoding block $D_m$ is reformed into two modules $U$ and $H$ as shown in Fig.~\ref{Fig:overview}: its $U$ module is an upsampling block operates on the output of the previous decoding block $D_{m+1}$ chaining till $I^{M-1}$ of the lowest resolution. $U$ in $D_m$ reverses the corresponding $E_m$ and replaces its last convolutional layer with a deconvolutional layer of stride 2 to upsample the image. 
The $H$ module is a 2-layer residual block~\cite{he2016cvpr} connecting to the mirrored encoding block $E_m$ for high-frequency communication. Assembling them we reconstruct the $m^\text{th}$ level lowpass image with higher resolution than the previous level ($m+1$). Analog to (\ref{eq:Lap}), the output $\tilde I^m$ of $D_m$ is obtained by,
\begin{equation}
\begin{split}
     \tilde I^m = U(\tilde I^{m+1}) + H(E_m).\\
\end{split}
\end{equation}
Different levels of details are manipulated in the encoding blocks through $H$: the high-frequency part of the image are gradually compressed from the frontend to the bottleneck of the network.}

\miaojing{The overall architecture is illustrated in Fig.~\ref{Fig:overview}. Similar to the hourglass structure in Fig.~\ref{Fig:AE}, we also adopt a stack of multiple decoding blocks denoted by $D_{M-1}$ to upsample the compressed representation from the bottleneck layer into the size of the $(M-1)^\text{th}$ scale lowpass image and associate the first ground truth $I^{M-1}$. These decoding blocks share the same filter size, stride and number of channels with that of $E_{M-1}$. Unlike in Fig.~\ref{Fig:AE}, the output channel of the last deconvolutional layer of $D_{M-1}$ in Fig.~\ref{Fig:overview} is reduced to 3 (or 1 for grey image) to directly produce the image $I^{M-1}$. Notice that it is important for $D_{M-1}$ to upsample the latent feature map into a reasonable size (\eg 7 $\times$ 7) before reducing its channels. The rest decoding blocks $D_m$ consist of $U$ and $H$ modules. Each $U$ module is an upsampling block, which is indeed similar to the naive decoding block in Fig.~\ref{Fig:AE}, but with its output channel being 3 (or 1). Each $H$ module is a 2-layer residual block consisting of two 3 $\times$ 3 convolutional layers and one skip connection: the number of channels of the first convoltuional layer is the same with its connected encoding layer, while the number of output channels of the second is set to 3 (or 1) for the same reason above; $1 \times 1$ convolutional layer is added on the skip connection to match the number of channels from input to output.}

\subsection{Loss function}\label{Sec:loss}
For image classification, we do not alter the training objective: the commonly used loss function is the cross-entropy loss denoted by $\mathcal L_{cls}$. For image reconstruction, we adopt the pixel-wise MSE loss between network prediction and ground truth. Referring to Sec.~\ref{Sec:joint}, the reconstruction loss is defined over the multi-scale output of the AE:
\begin{equation}\label{eq:recetloss}
\begin{split}
    \mathcal L_{rect} & = \mathcal L^0_{rect}  + \lambda \sum^{M-1}_{m=1} \mathcal L^m_{rect}\\
    & = \|\tilde I^0 -I^0\|^2_{2} + \lambda \sum^{M-1}_{m=1} \|\tilde I^m -I^m\|^2_{2}
\end{split}
\end{equation}
where
$\tilde I^0$ and $\tilde I^m$ are the multi-scale reconstruction from the AE (see Fig.~\ref{Fig:overview}).
Parameter $\lambda$ is the loss weight between the original scale ($I^0$) and other scales ($I^m$).
$M$ is the total number of scales which should not be too large as the lowest resolution would become too small to be sensible. In practice we make sure that no loss function is associated for output size smaller than {$7 \times 7$}. 

A multi-task loss function is defined over the image classification and reconstruction branches, as $ \mathcal L = \mathcal L_{cls} + \mathcal L_{rect}$. Notice that the AE and classification net do not necessarily share the weights of all the encoding layers, we will provide ablation study for this.

\miaojing{The network is in general trained in a \emph{joint learning} manner which is fast and effective. Nevertheless, on natural image dataset, joint learning from scratch can be hard particularly for the AE. In this context, we recommend an \emph{alternative learning} manner: the AE is trained on some epochs alone. Then, the weights of its encoder are copied into the encoding blocks of the classifier, which in turn is fine-tuned for some epochs.  Then, the weights of its encoding blocks are copied back to the encoder of the AE.
After a few cycles, both the AE and the classifier are on the right track, we can switch to the joint learning. } 
Inference of the classifier is disjointed from the AE (upper branch in Fig.~\ref{Fig:overview}).  



	\section{Experiments}\label{Sec:Experiment}
\subsection{Experimental Setup}\label{Sec:Dataset}
\para{Datasets.} Experiments are conducted on three popular benchmarks for adversarial learning: MNIST, CIFAR-10, and ImageNet. MNIST consists of 70,000 grayscale images of hand written digits, in which 60,000 of them are used for training and the remaining is for testing. CIFAR-10 consists of 60,000 32x32 colour images in 10 classes, with 6,000 images per class. There are 50,000 training images and 10,000 test images. As for ImageNet, we use the same 1000 test images from the NIPS 2017 adversarial defense challenge~\cite{liao2018cvpr}. 

\para{Implementation details.}\label{Sec:ExperimentalDetails}
We choose by default ResNet-18 with SGD optimiser trained on classification task. Its four encoding blocks are taken as the encoder for the AE in Defend++. The decoder design refers to Sec.~\ref{Sec:joint}. For MNIST, the initial learning rate is set to 0.01 and is decayed by a factor of 10 every 100 epochs until 250 epochs. The momentum is 0.9, weight decay is 0.0005, batch size is 256. For CIFAR-10 and ImageNet, the initial learning rate is set to 0.1 and is decayed by a factor of 10 every 100 epochs until 300 epochs. The momentum is 0.9, weight decay is 0.0005, and batch size is 256. 
Parameter $\lambda$ in (\ref{eq:recetloss}) is set to 0.01. 

\para{Adversarial examples.}\label{Sec:Adversarial}
Adversarial examples are generated from the given classification network via representative methods \ie FGSM~\cite{szegedy2014iclr}, BIM~\cite{kurakin2017iclr}, C\&W~\cite{carlini2017sp} and DDN~\cite{rony2019cvpr}.
For FGSM we set by default the magnitude $\epsilon$ of adversarial perturbations as 0.2, 0.05 and 0.01 for MNIST, CIFAR-10, and ImageNet, respectively; we also evaluate the performance of different $\epsilon$ in Sec.~\ref{Sec:discussion}. BIM and C\&W are iterative methods, we set the maximal iteration for BIM and C\&W to 100 and $5\times 20$, respectively. DDN is a SOTA attack, we use its public code to generate adversarial examples. 

\para{Adversarial defense and baseline.}\label{Sec:defense}
Adversarial training is a straightforward yet effective defense method. We adopt two schemes: one is adversarial training with FGSM examples~\cite{szegedy2014iclr} (denoted by {Adv. FGSM}) while the other is the proposed adversarial re-training with DDN~\cite{rony2019cvpr} (denoted by {Adv. DDN}). We also implement ComDefend~\cite{jia2019cvpr} using its published code. Our baseline is a classification network without any defense mechanism.  

\para{Evaluation protocol.}\label{Sec:protocol}
We evaluate the classification accuracy on both the original and adversarial sets.

\begin{table}[t]
	\centering
	\small
	\begin{tabular}{cccccc}
		\toprule 
{Defend++}  & FGSM & BIM & C\&W & DDN & Clean\\
\midrule
1 shared block & 84.78 & 53.09 & 98.95 & 98.97 & 99.42\\
 2 shared blocks  & 85.20 & 50.52 & 98.99 & 98.95 & 99.39\\
3	shared blocks  & 86.48 & 54.56 & \textbf{99.14} & \textbf{99.16} & 99.42\\
4	shared blocks  & \textbf{86.81} & \textbf{55.37} & 99.08 & 99.10 & \textbf{99.47}\\

\midrule
ours w/o MS-LP & 85.42 & 53.48 & 99.01 & 99.05 & 99.41\\
ours w/ MS-LI & 85.17 & 46.90 & 99.03 & 99.04 & 99.39\\
ours & \textbf{86.81} & \textbf{55.37} & \textbf{99.08} & \textbf{99.10} & \textbf{99.47}\\
\midrule
Hourglass &  {85.39} & {53.91} & {99.03} & {99.05} & {99.51} \\
Hourglass+SC & {84.40} & {51.61} & {90.00} & {99.10} & {99.42} \\
Hourglass+SA & {85.19} & {54.32} & {99.05} & {99.12} & {99.45} \\
\bottomrule

	\hline 
	\end{tabular}
	\caption{\small Ablation study of different components of Defend++ on MNIST dataset. Classification accuracy on adversarial and clean examples are reported against several adversarial attacks. \emph{ours} is equivalent to \emph{4 share blocks}. \emph{MS-LP} (multi-scale lowpass objective) is applied to all the hourglass structures.   }
	\label{Tab:Ablation}	
	\vspace{-0.6cm}
\end{table}

\subsection{MNIST}\label{Sec:Ablation}
We provide ablation study regarding the shared layers (ResNet blocks) between the classification net and AE, multi-scale low-pass objective and multi-scale high-frequency communication.  

\para {Shared layers between classification and AE.}
Recalling Sec.~\ref{Sec:loss}, the AE and classification net does not need to share all encoding layers.
Our default classification net (ResNet-18) has four encoding blocks. In Table~\ref{Tab:Ablation}, we ablate the number of their shared blocks from 1 to 4. It turns out sharing four blocks performs best; the classification accuracy on FGSM attack is 86.81 and on DDN is 99.10; the performance of 3 shared blocks performs very close to that of 4 shared blocks. 
For CIFAR-10 and ImageNet, the best performance occurs when sharing three blocks. If not specified, all experiments below follow this setting. 



\para{Multi-scale lowpass objective for AE.}
Table~\ref{Tab:Ablation} also presents our method without multi-scale low-pass objective (denoted by {ours w/o MS-LP}); the result \emph{ours} is indeed the same with that of 4 shared blocks. The proposed MS-LP clearly improves the defense accuracy on nearly every attack as well as the clean image set. To further investigate its effectiveness, we offer the result of using linear interpolation instead for multi-scale ground truth (denoted by ours w/ MS-LI). Its performance is in general inferior to ours. For instance, the accuracy is 85.17 \vs 86.81 on FGSM; 46.90 \vs 55.37 on BIM. 

\para{Multi-scale high-frequency communication for AE.} \miaojing{We compare to the basic hourglass structure in Fig.~\ref{Fig:AE}. The result in Table~\ref{Tab:Ablation} shows that Hourglass produces accuracy clearly lower than ours on every entry. As being suggested, the connection between encoder and decoder in the basic hourglass structure is only through the bottleneck, which is weak. In order to enhance this connection, we could add skip connections similar to U-Net~\cite{ronneberger2015miccai}, where corresponding encoding and decoding blocks (\eg $E^m$ and $D^m$) are connected using identity mappings. The feature tensors from two branches can be either concatenated or element-wise added, which we denote as Hourglass-SC and Hourglass-SA, respectively. Results in Table~\ref{Tab:Ablation} show no significant improvement over the basic Hourglass, and are even worse on some attacks (\eg 85.19 \vs 85.39 on FGSM). Hourglass-SA is slightly better than Hourglass-SC yet is still inferior to ours. Our network reformation offers more elaborate frequency-steering for encoding and decoding blocks, compared to the straightforward skip connection on Hourglass. Hence, it ends up with a more robust defense against various adversarial attacks.  }


\begin{table}[t]
	\centering
	\small
	\begin{tabular}{cccccc}
\toprule
		  & FGSM & BIM & C\&W & DDN & Clean\\
		\midrule
		No defense & 31.94 & 0.00 & 0.00 & 0.00 & \textbf{99.60} \\
	    Adv. FGSM & \textbf{99.77} & 9.86 & 99.04 & 99.05 & 99.51\\
		Adv. DDN & 95.21 & 19.55 & 98.66 & 98.66 & 98.94\\
		ComDefend & 85.69 & 19.04 & 98.87 & 98.82 & 99.43\\
		ComDefend-wb & 19.93 & 0.00 & 0.00 & 0.00 & 99.43\\
	ours  & 86.81 & {55.37} & \textbf{99.08} & \textbf{99.10} & 99.47\\
	ours + GN  & 90.56 & \textbf{56.44} & {98.97} &{98.98} & 99.41\\
\bottomrule
	\end{tabular}
	\caption{\small  Classification accuracy on adversarial and clean examples of selected defense methods. Experiment on MNIST dataset.}
	\label{Tab:MNIST}	
	\vspace{-0.6cm}
\end{table}

\begin{figure}[t]
	\centering
	\includegraphics[width=1\columnwidth]{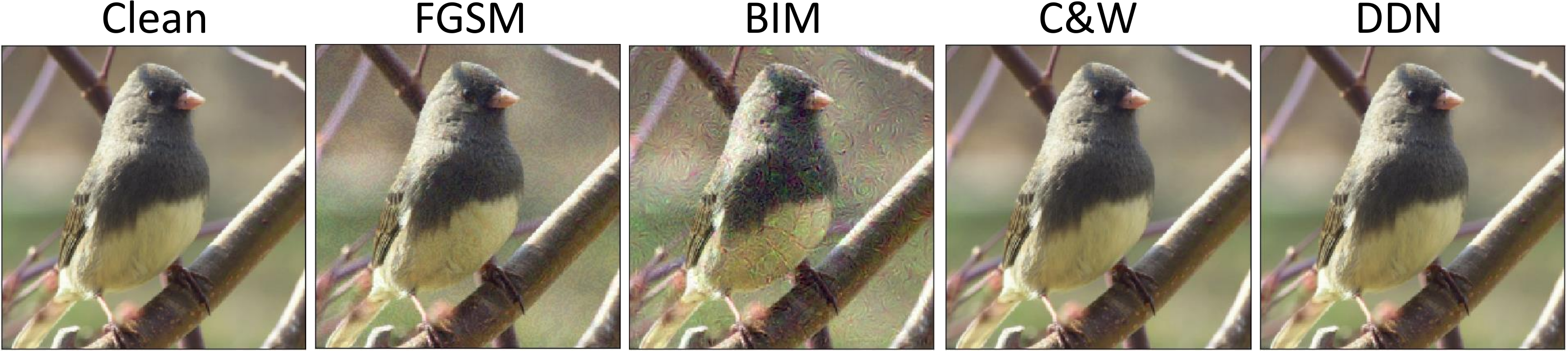}
\vspace{-6mm}
	\caption{Illustration of adversarial examples.    
	}
	\label{Fig:adversarial}
	\vspace{-0.3cm}
\end{figure}

\para{Comparison with representative defense methods.}
Table~\ref{Tab:MNIST} compares Defend++ to adversarial training with FGSM (Adv. FGSM), adversarial re-training with DDN (Adv. DDN)~\cite{rony2019cvpr}, ComDefend~\cite{jia2019cvpr} and baseline (No defense).
Defend++ significantly improves the performance of the baseline over any adversarial attack. The baseline and Defend++ are both trained with clean images only. Defend++ is inferior to Adv.~FGSM on the FGSM attack (86.81 \vs 99.77) because Adv.~FGSM is specifically trained for this attack. However, Defend++ shows a strong generalization ability compared to Adv. FGSM. For instance, it is 55.37 \vs 9.86 on BIM. BIM is the strongest attack which causes quite visible image distortion (see Fig.~\ref{Fig:adversarial}). Adv.DDN is indeed adversarial retraining where adversarial examples are online updated. This is very slow. More epochs are needed to obtain better performance~\cite{rony2019cvpr}. In contrast, Defend++ trains with clean images and is much faster. 


Performance of ComDefend is also reported under our setting, which performs close to Adv.~DDN but in a grey-box scenario: the attacker is oblivious to the reformer (AE) in front of the classification net~\cite{jia2019cvpr} (see Sec.~\ref{Sec:RelatedWorks}).
In a fair comparison with Defend++, where the reformer is known to the attacker in white-box scenario, ComDefend is no better than the baseline (see ComDefend-wb in Table~\ref{Tab:MNIST}). \miaojing{Our work is similar in spirit to ComDefend that we do not use any adversarial examples for training. Nonetheless, ComDefend adds Gaussian noises (GN) to image representations in the reformer to improve the defending performance. By doing the same, we can further obtain our results (see ours + GN): 90.56, 56.44, 98.97, 98.98 on FGSM, BIM, C\&W, DDN and clean examples, respectively; the accuracy is clearly improved by +3.75\% and +1.13\% on FGSM and BIM, while slightly declined on C\&W, DDN and clean examples.  
} 

\begin{table}[t]
	\centering
	\small
	\begin{tabular}{cccccc}
		\toprule
		  & FGSM & BIM & C\&W & DDN & Clean\\
\midrule 
		No defense & 38.35 & 0.00 & 0.00 & 0.00 & \textbf{93.80}\\
	Adv. FGSM & \textbf{74.01} & 11.68 & 89.63 & 89.46 & 91.30\\
	Adv. DDN  & 68.88 & 14.42 & 86.93 & 86.90 & 86.94\\
ComDefend & 58.06 & \textbf{23.60} & 77.41 & 76.28 & 87.58\\
	ours & 64.63 & {16.76} & {89.78} & \textbf{89.83} & 93.22\\
		ours + GN & 67.42 & {18.12} & \textbf{89.95} & 88.76 & 92.05\\
\bottomrule
	\end{tabular}
	\caption{\small Classification accuracy on CIFAR-10 dataset.}
	\label{Tab:CIFAR}	
	\vspace{-0.6cm}
\end{table}

\begin{table}[!t]
	\centering
	\small
	\begin{tabular}{cccccc}
		\toprule
		  & FGSM & BIM & C\&W & DDN & Clean\\
\midrule 
	No defense & 13.85 & 0.00 & 0.00 & 0.00 & \textbf{71.25}\\
	Adv. FGSM & \textbf{58.30} & 10.65 & 43.45 & 44.85 & 52.75\\
	ours & 56.45 & \textbf{40.85} & \textbf{54.85} & \textbf{56.35} & 62.45\\
\bottomrule
	\end{tabular}
	\caption{\small Classification accuracy on ImageNet dataset.}
	\label{Tab:ImageNet}	
	\vspace{-0.5cm}
\end{table}

\miaojing{Before the end, two more points are worth noting: 1) there exists a similar representative attack to BIM, the projected gradient 
descent (PGD)~\cite{madry2017arxiv}. They have the same multi-step generation process yet PGD uses uniform random noise as initialization. The testing results against PGD attack (100 iterations) on MNIST are indeed similar to BIM: 0.0, 5.12, 16.6, 15.24, 48.53 for No defense, Adv. FGSM, Adv. DDN, ComDefend, and our Defend++, respectively; Defend++ is clearly the best. On the other hand, for adversarial training with PGD (Adv. PGD) on MNIST, it is very slow and we got accuracy 83.62, 36.71, 97.97, 97.99, 61.14, 98.54 for FGSM, BIM, C\&W, DDN, PGD, clean examples, respectively; Defend++ is in general better than this Adv. PGD. 2) We notice that even those degraded versions of  Defend++, \eg ours w/o MS-LP or Hourglass in Table~\ref{Tab:MNIST}, perform much better than the baseline and competitive to other representative defense methods. This validates our idea in general of reinforcing the DNN bottleneck of a classifier with an AE for adversarial defense, which can significantly improve the defense \emph{universality}, without the need of adversarial training or pre-processing head. }

\subsection{CIFAR-10}
Table~\ref{Tab:CIFAR} presents the results on CIFAR-10. Like in Table~\ref{Tab:MNIST}, Defend++ significantly improves the defense performance over the baseline (No defense) on all the attacks. It is also clearly better than Adv. FGSM/DDN and ComDefend on C\&W and DDN attacks while inferior to Adv. FGSM on FGSM attack (64.63 \vs 74.01) and ComDefend on BIM attack (16.76 \vs 23.60). It maintains a relatively good accuracy on the clean set (93.22) while the others clearly drop (91.30, 86.94, and 87.58). Our Defend++ uses only clean images for training. Similar to that in Table~\ref{Tab:MNIST}, \teddy{we also add Gaussian noises (GN) during training as in~\cite{jia2019cvpr}: the results are further improved especially against simple attacks (FGSM and BIM) but the accuracy on clean images drops by one point (see Table~\ref{Tab:CIFAR}, ours + GN)}.

\subsection{ImageNet}
Results on ImageNet are reported in Table~\ref{Tab:ImageNet}. The No defense baseline demonstrates a total failure on BIM, C\&W, and DDN while our Defend++ substantially improves it. Defend++ also yields better result than Adv.FGSM on every attack except FGSM, where it is fairly lower as Adv.FGSM is trained with FGSM examples. Notice that experiments on ImageNet are only trained with a subset (around 10k images) of it for fast implementation so the overall accuracy is not very high. 
Fig.~\ref{Fig:adversarial} shows examples of attacked images. One can clearly see that BIM attack is more visible.


 \begin{figure}[t]
\centering
  \includegraphics[width=0.9\linewidth,height=3.5cm]{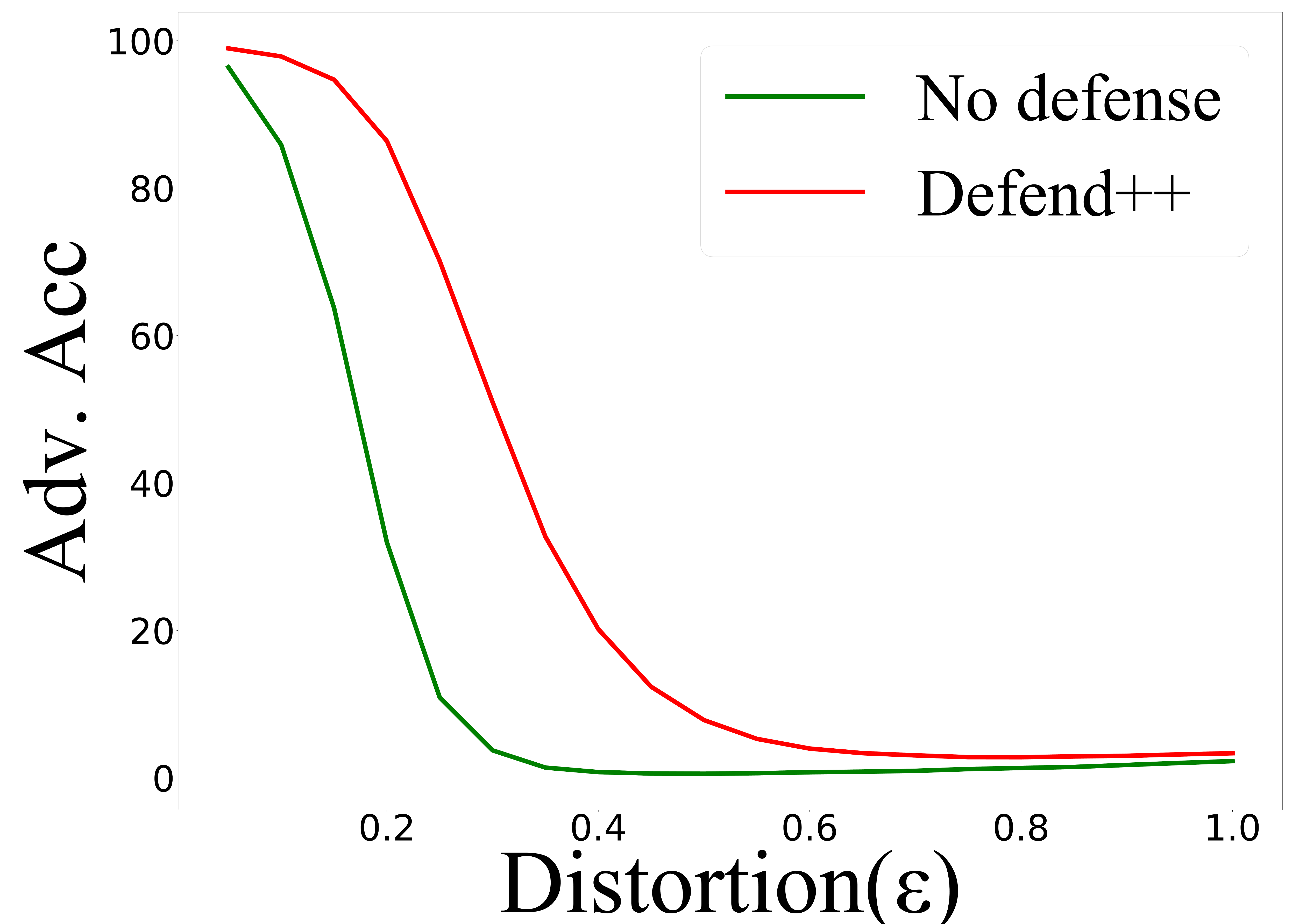}
  \vspace{-0.2cm}
  \caption{Defend++ against different distortions.}\label{fig:magnitude}
  \end{figure}

 \begin{figure}[t]
	\centering
	\includegraphics[width=0.9\columnwidth]{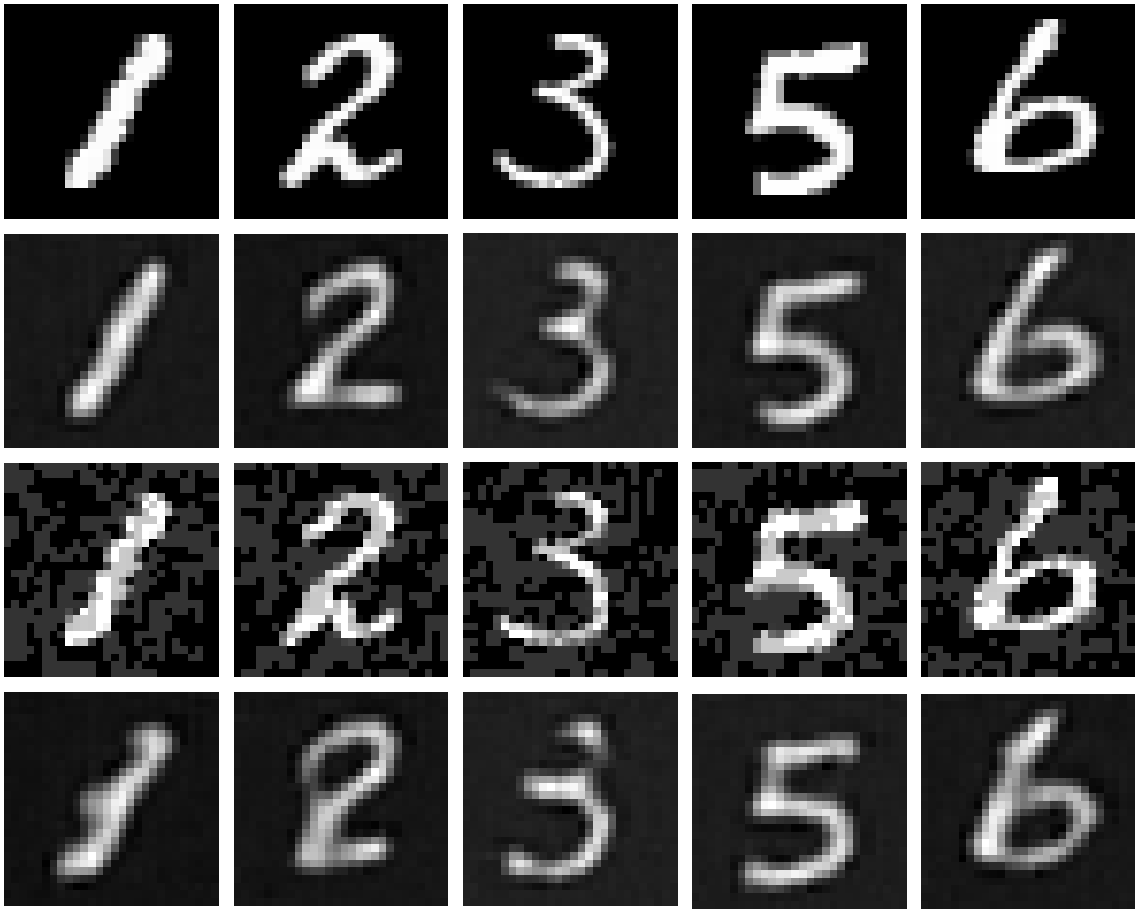}
	\vspace{-2mm}
	\caption{First and second rows: clean images and their reconstructed results via our AE. Third and fourth rows: adversarial images and their reconstructed results via our AE.   
	}
	\label{Fig:example}
	\vspace{-0.2cm}
\end{figure}


\begin{figure}[tb]
\centering
  \includegraphics[width=0.9\linewidth,height=3.5cm]{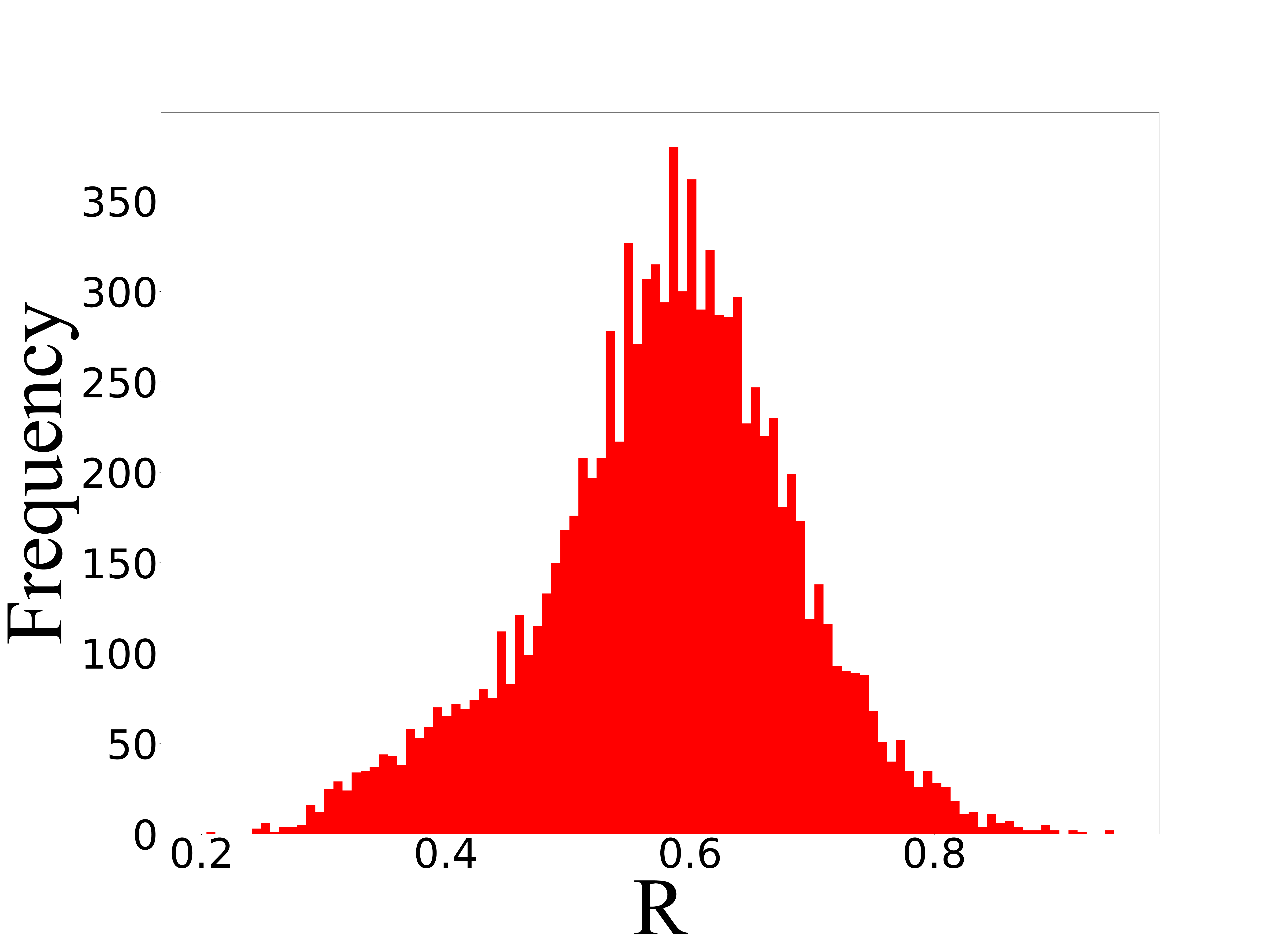}
  \vspace{-0.3cm}
  \caption{Frequency histogram of $R$.}\label{fig:distance}
 \vspace{-0.2cm}
\end{figure}

\subsection{Analysis and Discussion}\label{Sec:discussion}
We offer more analysis and discussion on MNIST and CIFAR-10. 


\para{Accuracy against various target distortions.}
We vary the magnitude $\epsilon$ of FGSM perturbations for Defend++ from 0.1 to 1.0. The accuracy on MNIST is illustrated in Fig.~\ref{fig:magnitude}.
By adopting Defend++, the baseline curve is substantially shifted to the right.
This clearly shows that our method improves the robustness of the classification network. The attacker has to almost double the distortion budget.

\begin{table}[!t]
	\centering
	\small
	\begin{tabular}{cccccc}
		\toprule
	MNIST & FGSM & BIM & C\&W & DDN & Clean\\
\midrule 
		No defense (VGG16) & 39.79 & 11.68 & 0.00 & 17.68 & \textbf{ 99.54}\\
	ours (VGG16) &\textbf{70.27} & \textbf{23.17} & \textbf{99.22} & \textbf{95.63} & 98.84\\
		\midrule
	No defense (ResNet-50) & 19.54 & 0.00 & 0.00 & 0.00 & \textbf{ 99.65}\\
	ours (ResNet-50) &\textbf{87.73} & \textbf{54.80} & \textbf{99.25} & \textbf{99.27} & 99.6\\
\bottomrule
	\end{tabular}
	\caption{\small Experiments on MNIST dataset with classification net being VGG16 and ResNet-50.}
	\label{Tab:VGG}	
	\vspace{-0.6cm}
\end{table}

\begin{table}[!t]
	\centering
	\small
	\begin{tabular}{cccccc}
		\toprule
	CIFAR-10 & FGSM & BIM & C\&W & DDN & Clean\\
\midrule 
		No defense (VGG16) & 35.24 & 7.23 & 0.00 & 0.00 & \textbf{ 88.25}\\
	ours (VGG16) &\textbf{56.81} & \textbf{21.35} & \textbf{81.93} & \textbf{81.84} & 86.60\\
		\midrule
	No defense (ResNet-50) & 23.45 & 0.00 & 0.00 & 0.00 & \textbf{93.97}\\
	ours (ResNet-50) &\textbf{79.18} & \textbf{48.76} & \textbf{90.52} & \textbf{90.59} & 93.85\\
\bottomrule
	\end{tabular}
	\caption{\small Experiments on CIFAR-10 dataset with classification net being VGG16 and ResNet-50.}
	\label{Tab:cifar-vgg}	
	\vspace{-0.5cm}
\end{table}

\para{Removal of adversarial perturbations.}
The whole AE is not part of the classification model, but for illustration purpose, we inspect its capability of removing adversarial noises (FGSM) on MNIST. The reconstructed images shown in Fig.~\ref{Fig:example} are much cleaner than the adversarial images. Quantitatively, we compute the Euclidean distances from the reconstructed image to its corresponding attacked image ($D_{ra}$) and to its clean image ($D_{rc}$).
Fig.~\ref{fig:distance} shows that the ratio $R = D_{rc}/D_{ra}$ always falls into the interval $(0,1)$: The reconstruction is always closer to the clean image than to the attacked one. This proves that the AE succeeds to filter out most of the adversarial perturbation. 


\begin{table}[t]
	\setlength{\tabcolsep}{0.2pt}
	\centering
	\small
		\begin{tabular}{@{\extracolsep{4pt}}c|cccccc@{}}
	\toprule
Classifier & \multicolumn{2}{c}{VGG16} & \multicolumn{2}{c}{ResNet-50} & \multicolumn{2}{c}{ResNet-18}\\
 \cline{2-3} \cline{4-5} \cline{6-7}

Attack & VGG16 & ResNet-18 & ResNet-50 & ResNet-18& ResNet-18 & VGG16 \\
\midrule 
		No defense  & 39.79 & 52.73 & 19.54 & 33.28 & 31.94 & 35.41\\
ours &\textbf{70.27} & \textbf{76.09} & \textbf{87.73} & \textbf{90.47} & \textbf{86.81} & \textbf{89.39}\\
\bottomrule
	\end{tabular}
	\caption{\small Evaluation of universality of Defend++ on MNIST.}
	\label{Tab:mnist-transfer}	
	\vspace{-0.6cm}
\end{table}

\begin{table}[t]
		\setlength{\tabcolsep}{0.2pt}
	\centering
	\small
		\begin{tabular}{@{\extracolsep{4pt}}c|cccccc@{}}
	\toprule
	Classifier & \multicolumn{2}{c}{VGG16} & \multicolumn{2}{c}{ResNet-50} & \multicolumn{2}{c}{ResNet-18}\\
		\cline{2-3} \cline{4-5} \cline{6-7}
Attack & VGG16 & ResNet-18 & ResNet-50 & ResNet-18 & ResNet-18 & VGG16 \\
\midrule 
		No defense  & 35.24 & 47.49 & 23.45 & 41.39 & 38.35 & 46.18\\
	ours &\textbf{56.81} & \textbf{64.22} & \textbf{79.18} & \textbf{86.13} & \textbf{64.63} & \textbf{75.45}\\
\bottomrule
	\end{tabular}
	\caption{\small Evaluation of universality of Defend++ on CIFAR-10.}
	\label{Tab:cifar-transfer}	
	\vspace{-0.9cm}
\end{table}

\para{Networks and universality.}
\miaojing{So far, we choose ResNet-18 as our classification net, but Defend++ can adapt any network with the same training procedure (see Sec.~\ref{Sec:ExperimentalDetails}).
Here, we select VGG16 and ResNet-50 as the alternative classification nets. For VGG16, we take its 13 convolutional layers as the encoder of AE, every two/three convolutional layers with one pooling layer in the VGGnet are taken as one block. We let AE share the first two block weights with the classification net. For ResNet-50, we take its first four encoding blocks as the encoder of the AE and share their encoding weights.  
The results are shown in Table~\ref{Tab:VGG} and ~\ref{Tab:cifar-vgg} on MNIST and CIFAR-10, respectively. Defend++ on VGG16 and ResNet-50 clearly improves the baseline on both adversarial and clean examples, which shows the generalizability of our method on different networks. Notice that 1) the results on ResNet-50 are also slightly better than those on ResNet-18 (Table~\ref{Tab:MNIST} and~\ref{Tab:CIFAR}), as ResNet-50 is a more powerful network; 2) the FGSM, BIM, C\&W results (79.18, 48.76, 90.52) on ResNet-50, CIFAR-10 are also comparable to ComDefend~\cite{jia2019cvpr} under similar setting (\eg 83, 34, 87 for $L_{\infty} = 16$ in its Table 5): ours performs better on BIM and C\&W attacks.} \teddy{More importantly, ours are pure white-box classifiers, contrary to ComDefend where the upfront reformer is not public (see Sec~\ref{Sec:Ablation})}.

\miaojing{Defend++ enhances the network defense against universal adversarial perturbations by removing them from the bottleneck representations. From this point of view, we expect that Defend++ can defend adversarial perturbations generated from different models. We use Defend++ trained on one network to defend against adversarial examples generated from another network. 
Results are reported on FGSM examples amid ResNet-18, ResNet-50 and VGG16 in Table~\ref{Tab:cifar-transfer} and~\ref{Tab:mnist-transfer}. 
Having a look at the tables, the first row signifies the classification network we use for testing while the second row signifies the classification network we use for adversary generation. For instance, given a classifier VGG16, its adversarial examples can be generated using the same VGG16 or using ResNet-18. 
Defend++ is not a \emph{gradient-masking} based approach like~\cite{papernot2016sp,buckman2018iclr,hua2019mm,nayebi2017arxiv}; the results show that Defend++ significantly improves the performance over baseline in the transfer setting, which demonstrates its strong universality.
Furthermore, comparing the cross-model performance with that using the same model, it is apparent that our Defend++ performs even better on the cross-model. Take the ResNet-50 classifier as an example, its accuracy on attacks from the same ResNet-50 is 87.73 and 79.18 on MNIST and CIFAR-10, respectively; on attacks from ResNet-18 is improved to 90.47 and 86.13 correspondingly. Adversarial examples generated from the same model with the classifier is harder to defend. }



	\section{Conclusion}
This paper proposes a DNN bottleneck reinforcement scheme for universal adversarial defense (Defend++). It learns an auto-encoder jointly with the classifier by sharing the same encoding parameters of the network. The auto-encoder improves the classifier's compressing ability by removing the adversarial/noisy perturbations at its encoding stage. Multi-scale low-pass objective and multi-scale high-frequency communication is also introduced at training to further improve the robustness of the network. Defend++ is trained with clean images only and without changing the classification structure.  Thorough experiments show that, compared to other representative methods, Defend++ is an effective defense against various  attacks at very low cost.
\pagebreak

\begin{acks}
This work was supported by the National Natural Science
Foundation of China (NSFC) under Grant No. 61828602 and 51475334; as well as National Key Research
and Development Program of Science and Technology of China under
Grant No. 2018YFB1305304, Shanghai Science and Technology Pilot
Project under Grant No. 19511132100, and the French ANR chair SAIDA. 
\end{acks}

\bibliographystyle{ACM-Reference-Format}
\bibliography{bib/cvpr2020}

\end{document}